\documentclass[10pt,letterpaper]{article}

\usepackage{ccn}
\usepackage{pslatex}
\usepackage{apacite}
\usepackage{graphicx}
\usepackage{multirow}
\usepackage{amsmath}

\title{From Images to Perception: \\ Emergence of Perceptual Properties by Reconstructing Images}

\author{{\large \bf Pablo Hernández-Cámara$^{*}$, \bf Jesus Malo, \bf Valero Laparra}\\
Image Processing Lab, University of Valencia \\
$^{*}$ Corresponding author: pablo.hernandez-camara@uv.es}

\begin{document}

\maketitle

\section{Abstract}
{
\bf
A number of scientists suggested that human visual perception may emerge from image statistics, shaping efficient neural representations in early vision. In this work, a bio-inspired architecture that can accommodate several known facts in the retina-V1 cortex, the PerceptNet, has been end-to-end optimized for different tasks related to image reconstruction: autoencoding, denoising, deblurring, and sparsity regularization.
Our results show that the encoder stage (V1-like layer) consistently exhibits the highest correlation with human perceptual judgments on image distortion despite not using perceptual information in the initialization or training. This alignment exhibits an optimum for moderate noise, blur and sparsity. 
These findings suggest that the visual system may be tuned to remove those particular levels of distortion with that level of sparsity and that biologically inspired models can learn perceptual metrics without human supervision.
}
\begin{quote}
\small
\textbf{Keywords:} 
Perceptual Representation; Visual Perception; Bio-Inspired Models; Self-Supervised Learning; Autoencoder
\end{quote}

\vspace{-0.2cm}
\section{Introduction}

On the one hand, following the classical Efficient Coding Hypothesis \cite{barlow1961possible, barlow2001redundancy}, many researchers have shown that certain behaviors of biological vision systems can be derived from the statistical regularities of natural images. Examples include: color coding based on one achromatic and two chromatic broad-band spectral sensitivities \cite{buchsbaum1983trichromacy} and their associated nonlinearities \cite{von2001optimal, laparra2012nonlinearities}; the emergence of achromatic and spatio-chromatic frequency analyzers 
\cite{olshausen1996emergence}, including their bandwidth \cite{atick1992understanding}, adaptation \cite{gutmann2014spatio}, and nonlinear responses \cite{schwartz2001natural}. On the other hand, autoencoders also capture the statistics of the images they are trained to reconstruct. Recent works have shown that denoising and deblurring autoencoders can reproduce the human contrast sensitivity function \cite{li2022contrast} and exhibit humanlike color illusions \cite{gomez2020color}. Similarly, compression autoencoders develop non-Euclidean metrics aligned with human judgments of image distortion \cite{hepburn2021relation}, while networks trained on low- and mid-level vision tasks also induce perceptually aligned distortion metrics \cite{kumar2022better, hernandez2025dissecting}.

These findings raise the question: Can biologically inspired architectures of the early visual system, such as PerceptNet \cite{hepburn2020perceptnet}, learn perceptual distances without explicit perceptual supervision?
In this work, we train PerceptNet on tasks including image reconstruction, denoising, deblurring, and sparsity regularization. We then analyze how these objectives influence the human alignment with perceptual judgments. Our results show that the strongest correlation with human evaluations arises at the V1 stage of PerceptNet. Notably, this alignment displays optimal values for moderate noise, blur, and sparsity levels.

By demonstrating that autoencoders can learn human-like perceptual properties, our study offers insights into both computational and neurobiological mechanisms of vision. Furthermore, it suggests that bio-inspired architectures may enable perceptual metrics that generalize across tasks without requiring perceptual human-labeled data.

\section{Methods}

We base our approach on PerceptNet, a biologically inspired model designed to mimic the visual pathway up to the primary visual cortex \cite{hepburn2020perceptnet}. To train the model in a self-supervised fashion, we implement an autoencoder architecture by using PerceptNet as an encoder and appending a PerceptNet inverse version as the decoder. This inverse model mirrors the original PerceptNet but replaces pooling operations with upsampling and divisive normalization with inverse divisive normalization, which performs multiplicative scaling instead of division. Both the encoder and decoder parameters are learned jointly during training.

We use approximately 200,000 natural images sampled from the ImageNet dataset to train the models. Each model is trained until convergence, with hyper-parameters adjusted for each objective to optimize performance. We train the model with four different objectives:
\vspace{-0.2cm}
\begin{itemize}
    \item Image reconstruction: Model trained to minimize the mean squared error (MSE) between input and reconstructed images.
    \vspace{-0.2cm}
    \item Denoising: Model trained to reconstruct the clean version of Gaussian noise-corrupted images, parameterized by the noise standard deviation ($\sigma$).
    \vspace{-0.2cm}
    \item Deblurring: Model trained to reconstruct the clean version of blurred images (by convolving them with a Gaussian kernel), parameterized by the standard deviation of the convolution kernel ($\sigma$).
    \vspace{-0.2cm}
    \item Sparsity: Model trained to reconstruct images while encouraging sparse representations by adding an $L_1$ penalty on the mean absolute value of the encoder activations. The sparsity is parameterized by scaling the $L_1$ penalty with a hyperparameter ($\lambda$).
    \vspace{-0.1cm}
\end{itemize}

\begin{figure*}[t]
\begin{center}
\includegraphics[width=0.98\textwidth]{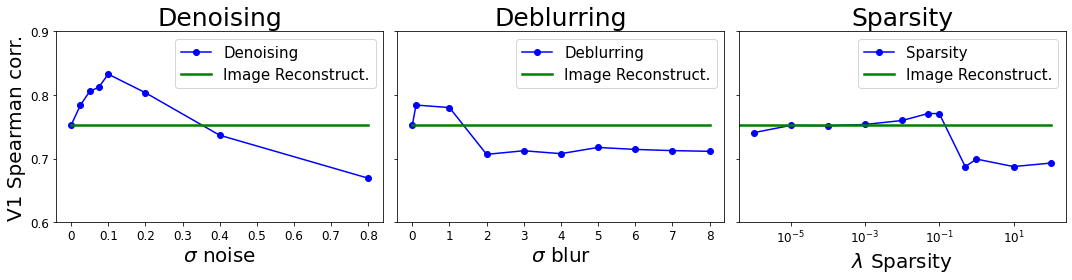}
\end{center}
\vspace{-0.75cm}
\caption{TID2013 Spearman correlation at the end of the encoder (V1-like layer) as a function of the different training parameters for denoising (left), deblurring (center) and sparsity (right).\vspace{-0.2cm}} 
\label{fig_all}
\end{figure*}

\begin{figure}[ht!]
\begin{center}
\vspace{-0.2cm}
\includegraphics[width=0.45\textwidth]{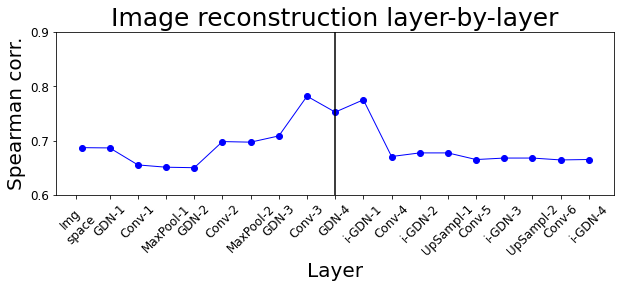}
\end{center}
\vspace{-0.7cm}
\caption{TID2013 Spearman correlation layer-by-layer when training the model to reconstruct natural images.\vspace{-0.45cm}} 
\label{fig_image_reconstruct}
\end{figure}

To evaluate perceptual alignment, we compute the correlation between differences in activations at each model layer and human Mean Opinion Score (MOS) from TID2013, a standard image quality assessment (IQA) database \cite{ponomarenko2015image}. This allows us to analyze how the different objectives affect the emergence of human-aligned perceptual representations at the different layers.

\vspace{-0.1cm}
\section{Results}

We first examine how the correlation with human MOS varies across the model's layers under the simpler image reconstruction objective. Figure \ref{fig_image_reconstruct} reveals a clear peak in correlation in the final layers of the encoder, corresponding to the V1-like stage of PerceptNet.

We then analyze the effect of different training objectives by focusing on the encoder output. Figure \ref{fig_all} left shows how the correlation increases with noise level, reaching a maximum at approximately $\sigma = 0.1$, after which it declines. This suggests that moderate noise levels improve perceptual alignment, but excessive noise reduces it.

Figure \ref{fig_all} center shows that a small blur with $\sigma \leq 1$ produces an increase in correlation, but a stronger blur reduces it. This suggests that the model benefits from learning to reverse slight degradations, but heavy blur impairs its ability to align with human perception.

Finally, figure \ref{fig_all} right shows that although sparsity has less effect than the previous goals, a moderate sparsity improves correlation, but higher levels ($\lambda > 0.1$) reduce it. This indicates a trade-off where sparsity enhances representations up to a point but reduces performance if over-enforced.

These findings suggest that perception emerges from efficient coding strategies, where the brain balances information preservation with noise suppression. The non-monotonic effects indicate that perception is optimized through an intermediate level of regularization rather than extreme constraints.

\vspace{-0.2cm}
\section{Conclusions}

Our study demonstrates that biologically inspired models can develop perceptual representations aligned with human vision through self-supervised learning alone, without perceptual supervision. Specifically, we show that PerceptNet, when trained as an autoencoder with appropriate regularization, exhibits emergent perceptual properties that strongly correlate with human judgments.

A key finding is that the highest alignment with human perception consistently arises at the encoder stage, which corresponds to V1 processing. This suggests that early visual representations in the brain may naturally reflect the statistical properties of the environment when optimized for reconstruction. Interestingly, moderate levels of noise, blur, and sparsity enhance this alignment, while excessive regularization reduces it. These results support that the visual system may be tuned to remove those particular levels of distortion with that level of sparsity and that biologically inspired models can learn perceptual metrics without human supervision.

Our findings complement and extend recent work showing that self-supervised models—such as denoising-deblurring autoencoders that replicate the human contrast sensitivity function \cite{li2022contrast} or compression autoencoders that learn human-aligned non-Euclidean metrics \cite{hepburn2021relation}—can capture perceptual properties without explicit supervision. Moreover, our analysis of task-dependent alignment patterns resonates with studies demonstrating that networks trained for low- and mid-level vision tasks can induce humanlike distortion metrics \cite{hernandez2025dissecting}.

By showing that biologically grounded architectures like PerceptNet can achieve similar alignment, our work provides further insight into the computational principles that may underlie biological vision. It also suggests that bio-inspired perceptual metrics could generalize across tasks and datasets, offering robust, interpretable models of perception without reliance on human annotations.

\section{Acknowledgments}

This work was supported in part by MCIN/AEI/FEDER/UE under Grants PID2020-118071GB-I00 and PID2023-152133NB-I00, by Spanish MIU under Grant FPU21/02256 and in part by Generalitat Valenciana under Projects GV/2021/074, CIPROM/2021/056, and by the grant BBVA Foundations of Science program: Maths, Stats, Comp. Sci. and AI (VIS4NN). Some computer resources were provided by Artemisa, funded by the EU ERDF through the Instituto de Física Corpuscular, IFIC (CSIC-UV).

\bibliographystyle{ccn_style}

\setlength{\bibleftmargin}{.125in}
\setlength{\bibindent}{-\bibleftmargin}

\bibliography{ccn_style}

\end{document}